\documentclass[10pt,twocolumn,letterpaper]{article}

\usepackage[pagenumbers]{cvpr} 

\definecolor{cvprblue}{rgb}{0.21,0.49,0.74}
\usepackage[pagebackref,breaklinks,colorlinks,allcolors=cvprblue]{hyperref}

\usepackage{booktabs}   
\usepackage{graphicx}   
\usepackage{multirow}   
\usepackage[table]{xcolor} 
\usepackage{amssymb}       
\usepackage{pifont}        

\usepackage{makecell}

\usepackage{tcolorbox}
\usepackage{graphicx}

\title{Video‑QTR: Query‑Driven Temporal Reasoning Framework for Lightweight Video Understanding}

\author{
  \textbf{Xinkui Zhao\textsuperscript{1,*}}, \quad
  \textbf{Zuxin Wang\textsuperscript{1}}, \quad
    \textbf{Yifan Zhang\textsuperscript{1}}, \quad
  \textbf{Guanjie Cheng\textsuperscript{2}}, \\
  \textbf{Yueshen Xu\textsuperscript{3}}, \quad
  \textbf{Shuiguang Deng\textsuperscript{2}}, \quad
   \textbf{Chang Liu\textsuperscript{2}}, \quad
    \textbf{Naibo Wang\textsuperscript{2}}, \quad
  \textbf{Jianwei Yin\textsuperscript{2}}
  \\
  \\
  \textsuperscript{1}School of Software Technology, Zhejiang University, Hangzhou, China \\
  \textsuperscript{2}School of Computer Science, Zhejiang University, Hangzhou, China \\
  \textsuperscript{3}School of Software Engineering, Xidian University, Xi'an, China \\
  \textsuperscript{*}Corresponding author.
}

\begin{document}
\maketitle


\begin{abstract}
The rapid development of multimodal large-language models (MLLMs) has significantly expanded the scope of visual language reasoning, enabling unified systems to interpret and describe complex visual content. However, applying these models to long-video understanding remains computationally intensive. Dense frame encoding generates excessive visual tokens, leading to high memory consumption, redundant computation, and limited scalability in real-world applications. This inefficiency highlights a key limitation of the traditional \textit{process-then-reason} paradigm, which analyzes visual streams exhaustively before semantic reasoning. To address this challenge, we introduce \textbf{Video-QTR (Query-Driven Temporal Reasoning)}, a lightweight framework that redefines video comprehension as a \textit{query-guided reasoning process}. Instead of encoding every frame, Video-QTR dynamically allocates perceptual resources based on the semantic intent of the query, creating an adaptive feedback loop between reasoning and perception. Extensive experiments across five benchmarks—MSVD-QA, ActivityNet-QA, MovieChat, and VideoMME—demonstrate that Video-QTR achieves \textbf{state-of-the-art performance} while reducing input frame consumption by up to \textbf{73\%}. These results confirm that query-driven temporal reasoning provides an efficient and scalable solution for video understanding.
\end{abstract}    

\section{Introduction}
\label{sec:intro}

\begin{figure}[t]
    \centering
    \includegraphics[width=1\linewidth]{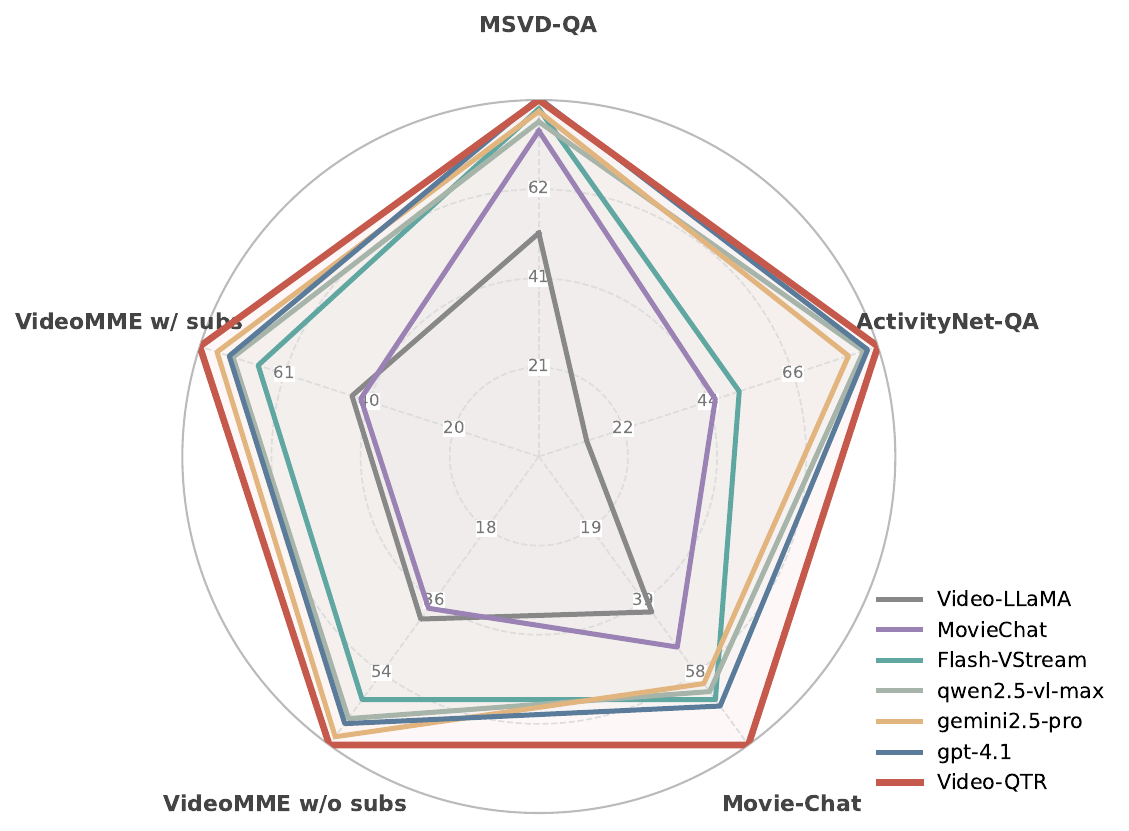}
    \caption{
    \textbf{Overall performance comparison with state-of-the-art models.} 
    We compare our method, \texttt{Video-QTR}, against leading video language models across five key benchmarks, covering both short video (MSVD-QA, ActivityNet-QA) and long video (MovieChat, VideoMME) understanding tasks.
    }
    \label{fig:overall_performance}
\end{figure}

The rapid advancement of Multimodal Large Language Models (MLLMs)~\cite{li2022blip,huang2023chatgpt,alayrac2022flamingo,li2023blip} has transformed how machines perceive and describe the visual world. MLLMs are now capable of tackling various vision-language tasks, such as chatbots, image captioning~\cite{vedantam2015cider,zhang2024differential}, and visual question answering~\cite{lu2022learn,liu2023visual,ye2023mplug}. These breakthroughs have paved the way toward general video comprehension, enabling models to analyze long, uncurated videos and generate structured, language-grounded responses, showing promise for embodied agents.

However, despite their success, current vision-language (VLA) pipelines are computationally intensive. Long videos consist of vast sequences of frames, each requiring significant processing power. As a result, the input may exceed MLLMs' context length limitations, and even with extended context lengths, processing long videos demands substantial computation and memory, posing challenges in real-world applications. Although several approaches have attempted to reduce the token count per frame~\cite{li2024llama,weng2024longvlm}, the typical "process-then-reason" approach still dominates.

From an information-theoretic perspective, visual streams exhibit low temporal entropy, while human cognition filters out irrelevant moments and focuses on goal-relevant segments. This mismatch raises an important question: \textit{Must a model process the entire visual stream to answer a video query?} Previous methods have attempted to reduce costs by highlighting salient frames or objects, assuming that visual importance directly correlates with reasoning relevance. However, saliency and relevance are not synonymous; visually prominent segments may be irrelevant to the query. We argue that effective video understanding should be \textbf{query-driven}, where perception and reasoning are dynamically guided by the query's intent, rather than by static visual prominence. Large Language Models (LLMs) have proven highly effective in structured reasoning and abstraction~\cite{brown2020language,achiam2023gpt,guo2025deepseek,liu2024deepseek}, and at a much lower computational cost than high-dimensional vision transformers. This disparity suggests a complementary design approach: while perception is essential, reasoning can be offloaded to the symbolic domain. By delegating temporal reasoning to the LLM, we reduce the computational load on the visual encoder. However, this transition introduces a fundamental challenge: \textbf{time}. Videos are inherently temporal, and understanding them requires reasoning \emph{through} time. In contrast, LLMs reason over symbolic sequences that lack continuous temporal flow, making it essential to bridge the gap between reasoning and temporal understanding.

To address these challenges, we introduce \textbf{Video-QTR} (\textbf{Q}uery-driven \textbf{T}emporal \textbf{R}easoning), a novel framework that redefines video comprehension. Instead of relying on exhaustive visual analysis, \textbf{Video-QTR} frames video understanding as an \textit{iterative reasoning process} guided by a \textit{language query}. The framework leverages an LLM to dynamically determine \textit{what} and \textit{when} to observe, creating a feedback loop between reasoning and perception.


\textbf{Video-QTR} consists of four interdependent components:
\begin{enumerate}
    \item \textbf{Reason Temporal Proxy (RTP):} Generates a temporal reasoning plan by analyzing the query and identifying relevant events and their time spans.
    \item \textbf{Perception Module:} Conducts on-demand visual grounding, extracting evidence from targeted segments to reduce computational overhead.
    \item \textbf{Temporal Consistency Refiner (TCR):} Aligns the inferred event sequence with the video timeline and refines reasoning.
    \item \textbf{Temporal Memory (TM):} Maintains a graph-like structure that updates temporal dependencies across iterations as reasoning evolves.
\end{enumerate}

We evaluate \textbf{Video-QTR} on four diverse benchmarks, covering short-form QA (MSVD-QA, ActivityNet-QA) and long-form comprehension (MovieChat, VideoMME). Despite avoiding dense frame encoding, \textbf{Video-QTR} achieves state-of-the-art performance across all settings, outperforming recent MLLMs such as Qwen2.5-VL-Max, Gemini 2.5 Pro, and GPT-4.1. Notably, on the challenging long-video benchmark \textit{MovieChat}, \textbf{Video-QTR} achieves 88.72\% accuracy in global reasoning, surpassing previous approaches by over 4 percentage points, while reducing visual token consumption by up to 73\%.

We summarize the key contributions of \textbf{Video-QTR}:
\begin{itemize}
    \item \textbf{Query-Driven Temporal Reasoning Paradigm:} We propose a shift from the conventional \textit{process-then-reason} paradigm to a \textit{reason-then-perceive} framework, where perception is guided by the query's semantic intent, eliminating redundant processing of task-irrelevant frames.
    \item \textbf{Lightweight, Iterative Video Reasoning Architecture:} \textbf{Video-QTR} introduces four synergistic components---RTP, Perception Module, TCR, and TM---that enable sparse, on-demand visual grounding while maintaining long-horizon temporal coherence.
    \item \textbf{Explicit Bridging of Symbolic Reasoning and Continuous Time:} We design the TCR module to correct the temporal order of LLM-generated reasoning plans against grounded visual evidence, ensuring chronological fidelity through iterative feedback.
    \item \textbf{State-of-the-Art Performance with High Efficiency:} \textbf{Video-QTR} achieves new state-of-the-art results on five video QA benchmarks, including both short-video (MSVD-QA, ActivityNet-QA) and long-video (MovieChat, VideoMME) settings, while significantly reducing visual token consumption (up to 73\% fewer tokens compared to dense encoding), demonstrating its practicality for real-world deployment.
\end{itemize}

\section{Related Work}
\subsection{Video-QA}
Video Question Answering (Video-QA) aims to answer questions about video content, requiring a comprehensive understanding of both visual and textual modalities. Datasets such as ActivityNet-QA~\cite{caba2015activitynet} and MSVD-QA~\cite{xu2017video} are commonly used for short-video understanding, while MovieChat~\cite{song2024moviechat} and VIDEO-MME~\cite{fu2025video} serve as benchmarks for long-video comprehension.

\begin{figure*}
    \centering
    \includegraphics[width=1\linewidth]{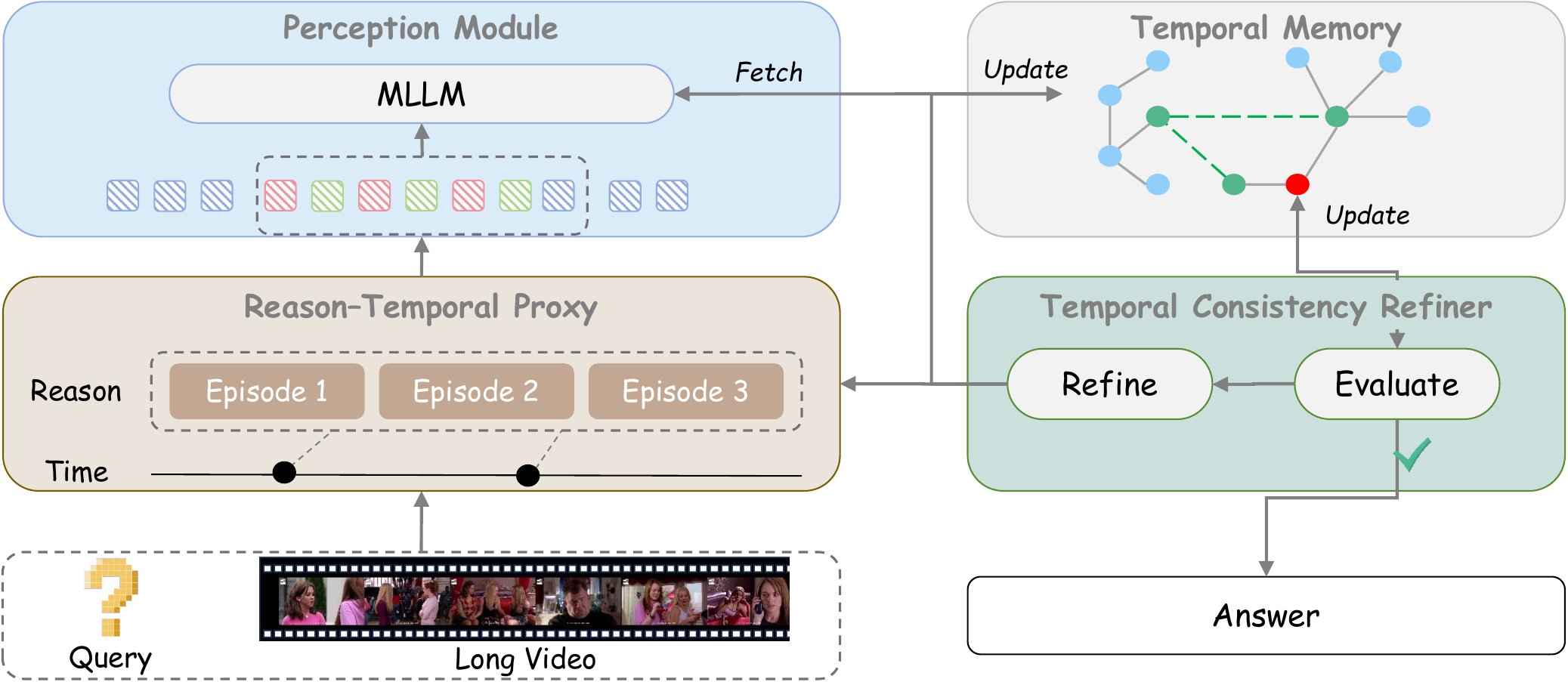}
    \caption{
\textbf{Framework overview.}
Video‑QTR performs query‑driven temporal reasoning through four cooperating modules.
The \textit{Reason–Temporal Proxy (RTP)} decomposes the query into temporal episodes along the video timeline.
The \textit{Perception Module} uses an MLLM backbone to selectively fetch visual evidence from relevant segments.
The \textit{Temporal Consistency Refiner (TCR)} evaluates and refines the chronological order between reasoning and observation.
The \textit{Temporal Memory (TM)} maintains an event graph that stores and updates semantic and temporal relations across iterations,
producing temporally consistent answers for long‑video understanding.
}
    \label{fig:placeholder}
\end{figure*}

\subsection{Long-Video Understanding}

Despite recent progress in multimodal large-language models (MLLMs)~\cite{li2022blip,alayrac2022flamingo,li2023blip} and their extensions to video understanding~\cite{jin2024chat,li2023videochat,li2024mvbench,liu2024st,luo2023valley,maaz2023video}, long-video understanding remains challenging due to the substantial temporal context and massive visual content involved.

A common approach is to expand the MLLM context window. Models such as MA-LMM~\cite{he2024ma}, SteamChat~\cite{xiong2025streaming}, and Movie-Chat~\cite{song2024moviechat} employ external memory modules for long-range reasoning. LWM~\cite{liu2024world} adopts RingAttention~\cite{liu2023blockwise} to enlarge the effective context, while MemViT~\cite{wu2021towards} and Compressed Memory Networks~\cite{wu2022memvit} compress memory representations to handle longer sequences efficiently.

Another research direction uses agent-driven workflows. DrVideo~\cite{ma2025drvideo} reformulates long-video tasks as long-document comprehension to exploit LLM reasoning. Socratic Models (SMs)~\cite{zeng2022socratic} enable heterogeneous models to communicate through natural language for zero-shot multimodal reasoning, and HERMES~\cite{faure2025hermes} provides modular components for enhancing or building long-form temporal reasoning systems.

Alternatively, some methods reduce input length by selecting key content. VideoLucy~\cite{zuo2025videolucy} employs a hierarchical memory with adaptive granularity and agent-based backtracking for global reasoning. VCA~\cite{yang2025vca} applies tree search to locate informative segments, while LVNet~\cite{park2024too} uses a Hierarchical Keyframe Selector (HKS) for keyframe-based captioning and question answering without full-video supervision. Selective token methods~\cite{wang2023selective} enhance long-form understanding via adaptive token selection and contrastive learning, and Tang et al.~\cite{tang2025adaptive} introduce an optimization-based keyframe selector that preserves informative frames under token constraints.

\section{Method}
\label{sec:method}

\subsection{Overview}
\label{sec:overview}

The \textbf{Video-QTR} (Query-driven Temporal Reasoning) architecture enables query-driven reasoning over long videos, balancing temporal coherence and computational efficiency. As shown in Fig.~\ref{fig:placeholder}, it consists of three core modules: \emph{Reason Temporal Proxy (RTP)}, \emph{Perception Module}, and \emph{Temporal Consistency Refiner (TCR)}, which work in an iterative, top-down flow from reasoning to perception. This design reframes video understanding as \textit{reasoning first, selective perception, and time-based verification}.

Given a query and video, the \textbf{RTP} generates a temporal reasoning plan to identify relevant segments and guide reasoning. The \textbf{Perception Module} encodes these segments into multi-scale visual representations, capturing key appearance and motion cues. The \textbf{TCR} evaluates the inferred sequence against the true timeline, refining reasoning and providing feedback.

This three-stage pipeline—\textit{temporal planning, visual grounding, and causal verification}—enables scalable video understanding without redundant frame processing. RTP ensures query-adaptivity, the Perception Module handles grounding, and the TCR maintains temporal fidelity. Sections~\ref{sec:rtp}, \ref{sec:tcr}, and \ref{sec:tm} describe each component in detail.

\begin{table*}[!ht]
    \centering
    \caption{
Performance comparison on \textbf{Video-MME} (long/overall durations) and \textbf{MovieChat} (Global, Breakpoint, Total modes). 
Abbreviations: \textbf{Long (-subs)} = long videos w/o subtitles, \textbf{Long (+subs)} = w/ subtitles, \textbf{Ovl (-subs)} = overall w/o subtitles, \textbf{Ovl (+subs)} = w/ subtitles, \textbf{Glob. Acc.} = global accuracy, \textbf{Glob. Scr.} = global score, \textbf{BP Acc.} = breakpoint accuracy, \textbf{BP Scr.} = breakpoint score.
}

    \label{tab:merged_performance}
    \resizebox{\textwidth}{!}{%
    \begin{tabular}{lcccccccc}
        \toprule
        & \multicolumn{4}{c}{\textbf{Video-MME}} & \multicolumn{4}{c}{\textbf{MovieChat}} \\
        \cmidrule(lr){2-5} \cmidrule(lr){6-9}
        \textbf{Method} 
        & \textbf{Long (-subs)} & \textbf{Long (+subs)} & \textbf{Ovl (-subs)} & \textbf{Ovl (+subs)} 
        & \textbf{Glob. Acc. (\%)} & \textbf{Glob. Scr.} & \textbf{BP Acc. (\%)} & \textbf{BP Scr.} \\
        \midrule
        
        \rowcolor{gray!20}
        \multicolumn{9}{c}{\textit{Comparison Methods}} \\
        Video-LLaVA~\cite{lin2023video} & 36.20 & 38.10 & 39.90 & 41.60 & 51.60 & 2.64 & 36.30 & 1.88 \\
        ST-LLM~\cite{liu2024st} & 31.30 & 36.90 & 37.90 & 42.30 & 59.81 & 2.96 & 46.20 & 2.33 \\
        VITA-1.0~\cite{fu2024vita} & 48.60 & 50.90 & 55.80 & 59.20 & - & - & - & - \\
        VideoChat2-Mistral~\cite{li2023videochat} & 33.20 & 39.20 & 39.50 & 43.80 & 44.62 & 2.23 & 43.03 & 2.18 \\
        VILA-1.5~\cite{fu2024vita} & \textbf{50.80} & 52.00 & 59.00 & 59.40 & - & - & 40 & 2.02 \\
        HERMES~\cite{faure2025hermes} & - & - & - & - & 78.60 & 3.99 & 57.30 & 2.88 \\
        VideoChat~\cite{li2023videochat} & 45.60 & 53.30 & 55.30 & 59.70 & 57.80 & 3.00 & 46.10 & 2.29 \\
        VideoLLaMA~\cite{li2024llama} & 47.60 & 49.00 & 52.40 & 54.71 & 51.70 & 2.67 & 39.10 & 2.04 \\
        Video-ChatGPT~\cite{maaz2023video} & - & - & - & - & 47.60 & 2.55 & 48.00 & 2.45 \\
        Flash-Vstream~\cite{zhang2024flash} & 50.30 & \textbf{61.40} & \textbf{61.20} & \textbf{67.00} & \textbf{86.00} & 4.28 & \textbf{59.60} & 3.01 \\
         MovieChat~\cite{song2024moviechat} & 33.40 & 36.13 & 33.40 & 37.11 & 62.30 & 3.23 & 48.30 & 2.57 \\

        \midrule
        \rowcolor{gray!20}
        \multicolumn{9}{c}{\textit{Large Multimodal Models}} \\
        gpt-4.1 & 50.51 & 61.04 & 67.23 & 73.90 & \textbf{87.63} & \textbf{4.38} & \textbf{61.39} & \textbf{3.10} \\
        gemini2.5-pro & \textbf{61.49} & \textbf{65.00} & \textbf{70.53} & \textbf{76.81} & 83.30 & 4.17 & 54.76 & 2.74 \\
        qwen2.5-vl-max & 51.47 & 60.37 & 65.96 & 72.99 & 86.70 & 4.33 & 56.59 & 2.82 \\

        \midrule
        \rowcolor{gray!20}
        \multicolumn{9}{c}{\textit{Our Method}} \\
        \textbf{Video-QTR} & \textbf{66.46} & \textbf{75.19} & \textbf{72.59} & \textbf{80.93} & \textbf{88.72} & \textbf{4.45} & \textbf{74.72} & \textbf{3.60} \\
        \bottomrule
    \end{tabular}%
    }
\end{table*}

\subsection{Reason Temporal Proxy}
\label{sec:rtp}

Traditional video-language models process videos with dense frame encoding, assuming full observation is needed to answer queries. While recent methods reduce costs by dynamically sampling frames or focusing on salient objects, they rely on visual prominence rather than reasoning relevance—what is visually prominent may not be relevant to the query. We propose a \textbf{query-driven} paradigm where perception adapts to the query’s intent, not static saliency. While perception is crucial, large-language models (LLMs) excel at abstraction and causal reasoning at a lower cost than vision encoders, allowing the LLM to control \emph{when} and \emph{where} to observe. The challenge is temporal grounding—language tokens are symbolic and discrete, while videos unfold continuously in time. To address this, we introduce the \textbf{Reason Temporal Proxy (RTP)}.

RTP allows the LLM to generate a temporal reasoning plan before frame processing. Given a query \( q \) and a video \( V \), RTP decomposes the query into temporal intents \(\{p_1, \dots, p_T\}\) with corresponding intervals \(\{\tau_1, \dots, \tau_T\}\). These intents guide perception, enabling reasoning-first and selective perception, reducing redundant computation and aligning linguistic reasoning with temporal causality.

\paragraph{Temporal Planning Mechanism.}
The LLM predicts event positions by generating reasoning tokens that encode semantic intent and temporal scope. Formally, the video is \( V = \{v_1, v_2, \dots, v_N\} \), where \( v_i \) is the \(i\)-th frame. RTP decomposes reasoning into \(T\) temporal episodes:
\[
\{(\langle tp_t \rangle, \tau_t)\}_{t=1}^{T}, \quad 
\tau_t = [s_t, e_t] \subseteq [1, N],
\tag{1}
\]
where each \(\langle tp_t \rangle\) is a \emph{temporal intent token} (i.e., \emph{what to look for}), and \(\tau_t\) indicates \emph{when} the event occurs. These intent-interval pairs serve as temporal anchors, guiding perception and improving interpretability.

\paragraph{Dynamic Reasoning Refinement.}
Static temporal splits can be suboptimal, as video relevance varies with the query. RTP refines temporal granularity based on semantic density and reasoning feedback. At each step,
\[
p_{t+1} = f_{\mathrm{LLM}}\!\left(q,\, \mathcal{M}_t,\, R_t\right),
\tag{2}
\]
where \(\mathcal{M}_t\) is the accumulated perception memory, and \(R_t\) is the reasoning state. The LLM refines the next temporal intent \(p_{t+1}\) based on hypothesis-evidence consistency. Dense transitions yield smaller intervals, while stable scenes are coarsely summarized. This iterative process aligns symbolic reasoning with temporal dynamics.

\subsection{Perception Module}
\label{sec:perception}

The \textbf{Perception Module} in Video-QTR performs \emph{query-conditioned selective perception}, in contrast to traditional encoders that process every frame uniformly. It receives temporal proposals from the Reason Temporal Proxy (RTP) and retrieves relevant visual evidence, linking symbolic reasoning with raw video signals. This reduces redundant computation and grounds reasoning in visual evidence.

\paragraph{Selective Video Grounding.}
Given a video \( V = \{v_1, v_2, \dots, v_N\} \) and temporal proposals \( \{\tau_1, \dots, \tau_T\} \) from RTP, the module retrieves a subset of frames:
\[
V \xrightarrow{\text{Select}} \widehat{V} = \{v_{i_1}, \dots, v_{i_K}\}, \qquad K \ll N.
\tag{1}
\]
Each frame \( v_{i_k} \) is encoded using a CLIP-ViT encoder to extract spatio-temporal features \(\mathbf{z}_{i_k}\), which are projected into the LLM input space:
\[
\mathbf{t}_{i_k} = W_2 \,\sigma(W_1 \mathbf{z}_{i_k}),
\tag{2}
\]
where \(\sigma(\cdot)\) is the GELU activation. The visual tokens \(\{\mathbf{t}_{i_k}\}\) are aligned with the symbolic intents \(\{p_1, \dots, p_T\}\) from RTP.

\paragraph{Contextual Integration.}
The retrieved features are integrated temporally to maintain consistency. For each intent \( p_t \) and segment \( \tau_t \), a context representation is computed:
\[
\mathbf{h}_t = \mathrm{Agg}\big(\{\mathbf{t}_{i_k} \mid v_{i_k} \in \tau_t \}\big),
\tag{3}
\]
where \(\mathrm{Agg}(\cdot)\) denotes temporal averaging or transformer aggregation. The aggregated vectors \(\{\mathbf{h}_t\}\) are passed to the LLM as visual tokens, grounding each reasoning step in the relevant context.

\paragraph{Adaptive Perception Strategy.}
The module adapts the sampling rate based on semantic variance, using denser sampling for high-motion or appearance changes and sparser sampling for stable regions, ensuring high-quality evidence without increasing computational cost.

\subsection{Temporal Consistency Refiner (TCR)}
\label{sec:tcr}

The Reason Temporal Proxy (RTP) generates a query-driven temporal plan, but its predicted sequence may not always align with the video’s timeline. The \textbf{Temporal Consistency Refiner (TCR)} ensures temporal consistency by refining the reasoning trajectory based on visual evidence.

\paragraph{Temporal Verification.}
For each reasoning cue \( r_t \) from RTP and its corresponding visual feature \( v_t \), TCR measures temporal coherence via cosine similarity:
\[
C_t(i) = \mathrm{sim}(E_q(r_t), \phi_i),
\tag{1}
\]
where \( E_q(\cdot) \) is the language embedding. The alignment distribution \( P_t(i) = \mathrm{Softmax}(C_t(i)) \) indicates the likelihood of matching temporal position \( i \) in the video.

\paragraph{Temporal Refinement.}
TCR refines reasoning by minimizing the difference between the expected and observed temporal positions:
\[
\mathcal{L}_{\mathrm{tcr}} = \frac{1}{T} \sum_{t=1}^{T} \Big\|\; P_t - \mathrm{Target}(\tau^{\mathrm{plan}}_t) \Big\|_2,
\tag{2}
\]
where \( \mathrm{Target}(\tau^{\mathrm{plan}}_t) \) is the one-hot target based on RTP’s predicted position. The gradients are backpropagated to improve future reasoning plans.

\paragraph{Iterative Feedback Loop.}
TCR establishes an iterative planning-verification loop:
\[
r_t \rightarrow F_t \rightarrow v_t \xrightarrow{\text{TCR}} (\mathcal{L}_{\mathrm{tcr}}, M_t) \rightarrow r_{t+1}.
\tag{3}
\]
Each iteration updates the Temporal Memory \( M_t \), allowing self-correction of temporal reasoning over long sequences.

\begin{table}[t!]
    \centering
    \caption{Performance comparison on ActivityNet-QA and MSVD-QA. Accuracy (Acc.) is reported in percentage (\%).}
    \label{tab:vqa_benchmark}
    \resizebox{\linewidth}{!}{%

    \begin{tabular}{lcccc}
        \toprule
        \multirow{2}{*}{\textbf{Method}} & \multicolumn{2}{c}{\textbf{ActivityNet-QA}} & \multicolumn{2}{c}{\textbf{MSVD-QA}} \\
        \cmidrule(lr){2-3} \cmidrule(lr){4-5}
         & \textbf{Acc. (\%)} & \textbf{Score} & \textbf{Acc. (\%)} & \textbf{Score} \\
        \midrule
        \rowcolor{gray!20}
        \multicolumn{5}{c}{\textit{Video QA Methods}} \\
        MovieChat\cite{song2024moviechat} & 45.70 & 3.40 & 75.20 & 3.80 \\
        VideoChat\cite{li2023videochat} & 26.50 & 2.20 & 56.30 & 2.80 \\
        VideoLLaMA\cite{li2024llama} & 12.40 & 1.10 & 51.60 & 2.50 \\
        Video-ChatGPT\cite{maaz2023video} & 35.20 & 2.70 & 64.90 & 3.30 \\
        PLLaVA\cite{xu2024pllava} & 56.30 & 3.50 & 76.60 & 4.10 \\
        IG-VLM\cite{kim2024image} & \textbf{57.30} & 3.60 & 76.70 & 4.10 \\
        Flash-Vstream\cite{zhang2024flash} & 51.90 & 3.40 & \textbf{80.30} & 3.90 \\
        \midrule
        \rowcolor{gray!20}
        \multicolumn{5}{c}{\textit{Large Multimodal Models}} \\
        qwen2.5-vl-max & 77.32 & 4.01 & 84.14 & 4.51 \\
        gemini2.5-pro & 79.70 & 3.88 & 80.20 & 3.90 \\
        gpt-4.1 & \textbf{82.78} & 4.13 & \textbf{85.15} & 4.56 \\
        \midrule

        \rowcolor{gray!20}
        \multicolumn{5}{c}{\textit{Our Method}} \\
        \textbf{Video-QTR} & \textbf{82.32} & \textbf{4.31} & \textbf{87.80} & \textbf{4.63} \\
        \bottomrule
    \end{tabular}
    }
\end{table}

\subsection{Temporal Memory (TM)}
\label{sec:tm}

To handle long-horizon video understanding, we introduce \textbf{Temporal Memory (TM)}, a \emph{graph-based memory} that encodes the video as a dynamic event graph anchored to time indices. The graph structure offers flexibility in updating event details and explicitly preserves temporal dependencies. This allows Video-QTR to accumulate reasoning states and maintain semantic and chronological continuity across iterations.

\paragraph{Graph-based Temporal Representation.}
At step \( t \), the memory is represented as \( \mathcal{G}_{t-1} = (\mathcal{V}_{t-1}, \mathcal{E}_{t-1}) \), where each node \( m_i \in \mathcal{V}_{t-1} \) corresponds to an event hypothesis with a temporal anchor \( \tau_i = [s_i, e_i] \) and stores a multimodal embedding. Edges \( (m_i, m_j, \alpha_{ij}) \in \mathcal{E}_{t-1} \) encode temporal relations (e.g., \texttt{before}, \texttt{after}) and causal dependencies, with confidence \( \alpha_{ij} \). TM updates the graph using the current perception \( \tilde{r}_t \) and corrective signal \( \Delta_t \):
\[
\mathcal{G}_t = \mathrm{UpdateGraph}\big(\mathcal{G}_{t-1}, \tilde{r}_t, \Delta_t\big).
\tag{1}
\]
New event nodes are added for unseen intervals, or existing nodes are refined when \( \tau_i \) overlaps with new evidence. Temporal edges are re-weighted based on TCR alignment, evolving the memory into a causally-connected graph.

\paragraph{Cross-Iteration Temporal Reasoning.}
The updated graph \( \mathcal{G}_t \) serves as temporal context for subsequent reasoning:
\[
r_{t+1} = f_{\mathrm{RTP}}(Q, \mathcal{G}_t), \qquad
\Delta_{t+1} = f_{\mathrm{TCR}}(\tilde{r}_{t+1}, \mathcal{G}_t).
\tag{2}
\]
Through message passing along temporal edges, the model retrieves past events and propagates information across time intervals, inferring higher-order relationships, such as cause-effect or recurring patterns, unifying localized episodes into a global trajectory. This enables reasoning beyond the perceptually visible window.

\section{Experiment}
\label{exp}

\subsection{Benchmarks}
We evaluate Video-QTR on representative Video-QA benchmarks, covering both short- and long-video understanding.

\begin{itemize}
    \item \textbf{ActivityNet-QA}~\cite{caba2015activitynet}: A large-scale dataset of human activity videos with question-answer pairs, evaluating models' understanding of real-world actions and events.
    \item \textbf{MSVD-QA}~\cite{xu2017video}: A dataset based on the Microsoft Research Video Description Corpus, focusing on fine-grained comprehension of short-video segments.
    \item \textbf{MovieChat}~\cite{song2024moviechat}: A large-scale long-form video dataset from movies and TV series, requiring reasoning across multi-scene and long temporal contexts.
    \item \textbf{VIDEO-MME}~\cite{fu2025video}: A benchmark covering videos of varying lengths and genres.
\end{itemize}

These benchmarks cover varied temporal lengths and reasoning complexities, enabling a thorough evaluation of Video-QTR’s performance on both short and long videos.

\begin{figure}[t]
    \centering
    \includegraphics[width=1\linewidth]{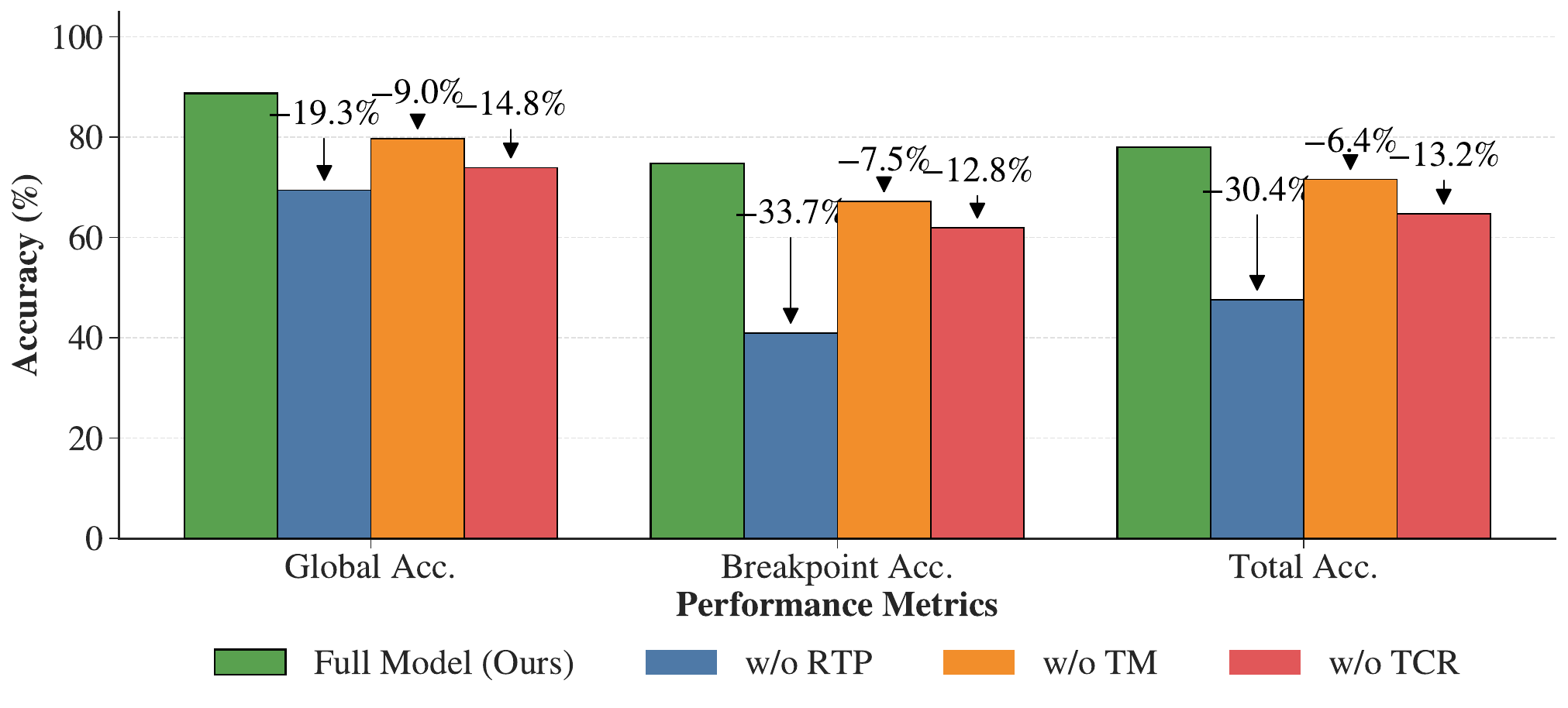}
    \caption{\textbf{Ablation study on MovieChat benchmark}. RTP and TCR significantly impact performance, while TM enhances long horizon stability and contextual reasoning.}

    \label{fig:aba}
\end{figure}
\subsection{Main Results}

\begin{figure*}[htbp]
    \centering
    \includegraphics[width=\textwidth]{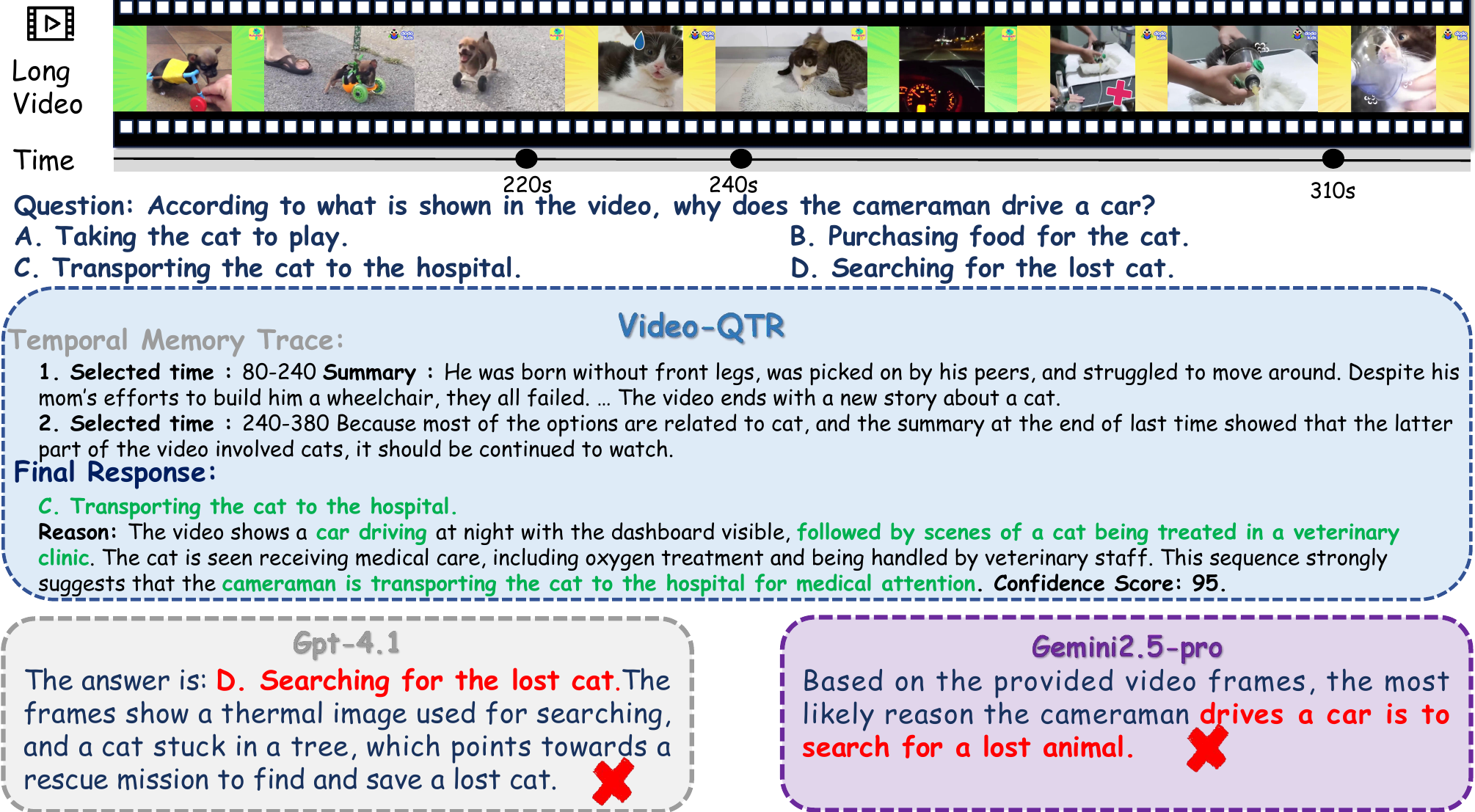} 
    \caption{\textbf{Qualitative Comparison of \texttt{Video-QTR} on Long Video Understanding.} Our \texttt{Video-QTR} demonstrates superior temporal reasoning and accurate detail perception, outperforming leading LLMs.}
    \label{fig:case_study}
\end{figure*}

Table~\ref{tab:merged_performance} presents the performance of our model on two long-video benchmarks, \textbf{Video-MME} and \textbf{MovieChat}, where it outperforms existing methods and large multimodal models. On \textbf{Video-MME}, we achieve state-of-the-art results with \textbf{72.59\%} and \textbf{80.93\%} accuracy for \textit{without-} and \textit{with-subtitle} settings, respectively, and excel in the \textbf{long-duration} scenario with \textbf{66.46\%} and \textbf{75.19\%} accuracy. Subtitles enhance the model's ability to capture fine-grained temporal and semantic cues, demonstrating the effectiveness of query-driven reasoning.

On \textbf{MovieChat}, we achieve \textbf{88.72\%} accuracy (4.45 score) in \textit{Global} mode and \textbf{74.72\%} accuracy (3.60 score) in \textit{Breakpoint} mode, outperforming both open-source and proprietary models.

Additionally, on short-video benchmarks \textbf{ActivityNet-QA} and \textbf{MSVD-QA} (Table~\ref{tab:vqa_benchmark}), our model achieves \textbf{82.32\%} (4.31 score) and \textbf{87.80\%} (4.63 score) accuracy, respectively, surpassing all existing Video-QA frameworks and competing with \textbf{GPT-4.1}. These results demonstrate that Video-QTR excels in both long- and short-video reasoning, providing robust answers in diverse temporal contexts.

\begin{table}[htbp]
\centering
\caption{
Performance comparison on Movie-Chat. 
\textbf{Bold} and \underline{underlined} denote the best and second best results, in order. 
Asterisk (*) indicates average frames used by our on demand process.
}
\label{tab:moviechat_comparison}
\begin{tabular}{lccc}
\toprule
\textbf{Method} & \textbf{Frames} & \makecell{\textbf{Global} \\ \textbf{Mode}} & \makecell{\textbf{Breakpoint} \\ \textbf{Mode}} \\
\midrule
MovieChat+\cite{song2025moviechat+} & 2048 & 71.20 & 49.60 \\
VideoChat\cite{li2023videochat} & 64 & 57.80 & 46.11 \\
VideoLLaMA\cite{li2024llama} & 32 & 51.70 & 39.10 \\
Video-ChatGPT\cite{maaz2023video} & 100 & 47.60 & 48.10 \\
mPLUG-2\cite{xu2023mplug} & 8 & 31.70 & 30.81 \\
\midrule
qwen2.5-vl-max & 512 & \underline{86.70} & \underline{56.59} \\
gemini2.5-pro & 1024 & 83.30 & 54.76 \\
\midrule
\textbf{Video-QTR} & 202.39* & \textbf{88.72} & \textbf{74.72} \\
\bottomrule
\end{tabular}
\end{table}

\subsection{Qualitative Analysis}
\label{sec:qualitative_analysis}

While the quantitative metrics in Section~\ref{sec:qualitative_analysis} demonstrate the superior performance of \texttt{Video-QTR}, qualitative analysis provides deeper insights into its query-driven reasoning, particularly for complex long-form video tasks where other models struggle.

We illustrate this with a case study in Figure~\ref{fig:case_study}, where the question is: "Why does the cameraman drive a car?" This task requires precise temporal grounding to distinguish relevant events from distractions. The video initially features a disabled dog (20-240s), before shifting to a narrative involving a cat. As shown in Figure~\ref{fig:case_study}, \texttt{Video-QTR} correctly identifies the answer, \textbf{C. Transporting the cat to the hospital}, with a confidence score of 95. The \textbf{Temporal Memory Trace} demonstrates its iterative reasoning process. In the first step, the model processes the segment from 80 to 240 seconds, noting the transition from the dog to the cat narrative. In the second step, guided by the query, it analyzes later segments (240-380s), identifying key events: a car driving at night and the cat receiving treatment. This reasoning, stored in \texttt{Temporal Memory}, leads to the correct identification of the cameraman's purpose.

In contrast, leading models like \texttt{GPT-4.1} and \texttt{Gemini2.5-pro} incorrectly suggest \textbf{D. Searching for the lost cat}, referencing hallucinated details like thermal imagery or a cat stuck in a tree. This highlights the limitation of traditional models, which often rely on exhaustive processing without query-driven guidance, leading to confusion in complex narratives. Our findings reinforce the quantitative results, showcasing \texttt{Video-QTR}'s superior ability for focused, context-aware, and temporally precise video understanding.



\subsection{Ablation Study}

We perform ablation studies in Figure~\ref{fig:aba} to evaluate the contribution of each core component in Video-QTR: Reason–Temporal Proxy (RTP), Temporal Memory (TM), and Temporal Consistency Refiner (TCR).

\paragraph{Reason–Temporal Proxy}



To assess the role of query-driven temporal planning, we replace RTP with a random segment selector, removing the model’s adaptive reasoning capability. As shown in Table~\ref{fig:aba}, this leads to a 30.4\% drop in total accuracy, underscoring RTP's importance in aligning symbolic reasoning with temporal events. This confirms that query-conditioned temporal planning is crucial for efficient long-video understanding.

\paragraph{Temporal Consistency Refiner}
We evaluate the effect of removing TCR, which eliminates the feedback loop enforcing temporal causality. This results in a 13.25\% performance drop, highlighting the necessity of consistency checking between language reasoning and temporal flow.

\paragraph{Temporal Memory}
Disabling Temporal Memory, which aggregates information across iterations, causes a 6.35\% accuracy decline, demonstrating the value of long-term contextual accumulation in cross-scene and cross-event reasoning. TM stabilizes inference and enables progressive comprehension across extended video sequences.

\subsection{Cost Analysis}
\label{sec:cost_analysis}

In addition to achieving state-of-the-art accuracy, a key contribution of \texttt{Video-QTR} is its computational efficiency. We compare the average number of frames processed by each model on the Movie-Chat benchmark, as shown in Table~\ref{tab:moviechat_comparison}. This metric reflects the computational and memory overhead for video understanding.

\texttt{Video-QTR} sets a new standard in both performance and efficiency. It achieves the highest scores in Global Mode (88.72\%) and Breakpoint Mode (74.72\%), processing only \textbf{202.39 frames} on average. These frames are selectively sampled via our \textit{on-demand} reasoning process, rather than through uniform pre-processing, which is central to the model’s lightweight design.

In contrast, \texttt{qwen2.5-vl-max}, the second-best model, processes 512 frames—over \textbf{2.5 times} more than our approach. \texttt{Gemini2.5-pro} requires 1024 frames, about \textbf{5$\times$} more, yet still underperforms. The gap is even wider compared to methods like \texttt{MovieChat+}, which processes 2048 frames—over \textbf{10$\times$} more than our model—with significantly lower accuracy.

While methods like \texttt{mPLUG-2} are extremely sparse (processing only 8 frames), they suffer from significant performance degradation. In contrast, \texttt{Video-QTR} strikes a balance, using its query-driven mechanism to process only the most relevant temporal segments, maintaining both efficiency and high performance.

These results empirically \textbf{validate our core design philosophy} outlined in Section~\ref{sec:method}. By offloading high-level temporal planning to the LLM and selectively invoking perception, \texttt{Video-QTR} navigates the efficiency-accuracy tradeoff more effectively than existing methods. This shows that comprehensive video understanding does not require exhaustive frame processing, enabling more scalable and practical systems for long-video analysis.

\begin{figure}[t]
    \centering 
    \includegraphics[width=\columnwidth]{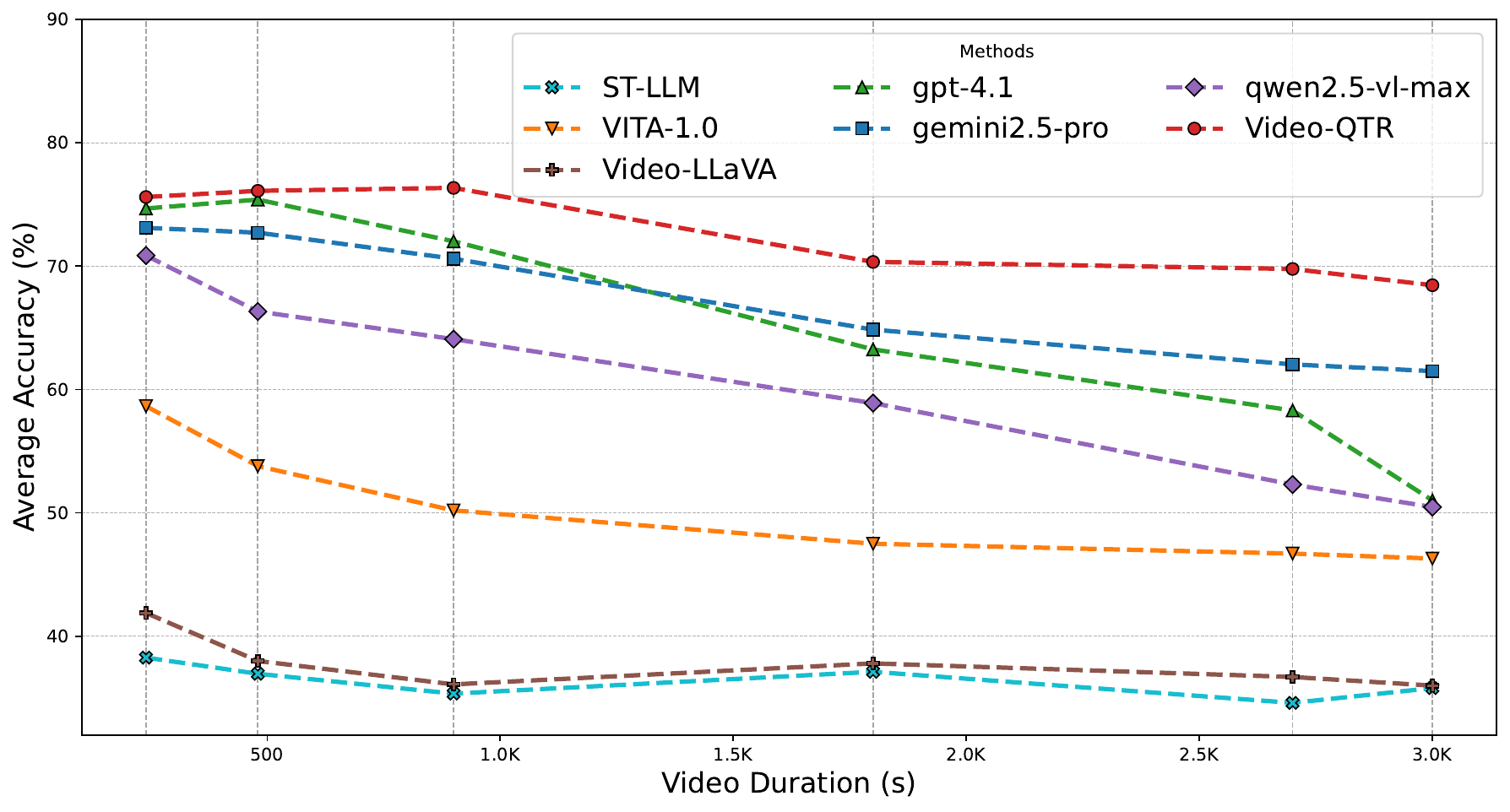}
    \caption{Performance comparison across video durations. Video-QTR outperforms all methods at every evaluation point, with accuracy decreasing for most models as video duration increases.}
    \label{fig:model_performance_duration} 
\end{figure}

\begin{figure}[t]
    \centering 
    \includegraphics[width=\columnwidth]{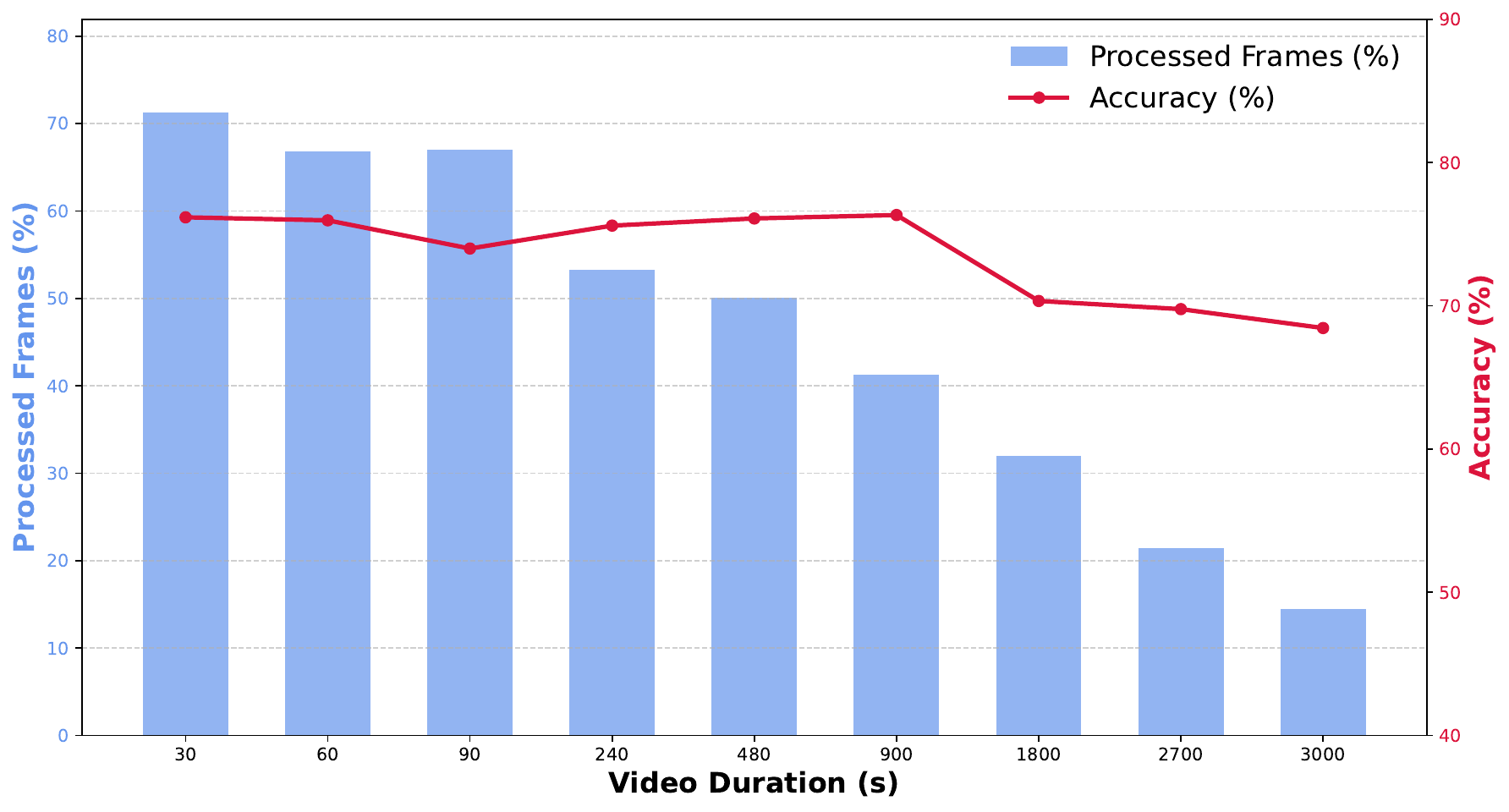}
    \caption{
\textbf{Efficiency and accuracy of \texttt{Video-QTR} across video durations.} 
Bars (left axis) show processed frames, while the line (right axis) shows accuracy.
}
\label{fig:qtr_efficiency_accuracy}
\end{figure}
\subsection{Performance-Cost Trade-off Analysis}
\label{sec:tradeoff_analysis}

A central challenge in long-video understanding is balancing computational cost with performance. Existing models often exhibit degraded accuracy and increased resource consumption as video duration extends, primarily due to exhaustive processing of redundant visual information. In contrast, \texttt{Video-QTR} redefines this balance.

As shown in Figure~\ref{fig:model_performance_duration}, \texttt{Video-QTR} consistently maintains the highest accuracy across all video lengths, demonstrating superior robustness even under significantly longer durations. This sustained performance is achieved with a remarkably lower computational budget, attributable to its query-driven temporal reasoning paradigm. Figure~\ref{fig:qtr_efficiency_accuracy} further illustrates this efficiency. The bar chart (left axis) reveals that \texttt{Video-QTR}'s computational cost, measured by the percentage of frames processed, decreases significantly with video length. For instance, it processes approximately 71\% of frames for a 30-second video, but only 15\% for a 3000-second video, showcasing adaptive scalability. The line plot (right axis) confirms that this substantial reduction in processed frames does not compromise performance; \texttt{Video-QTR} consistently outperforms other methods.

In conclusion, this analysis empirically validates that \texttt{Video-QTR} effectively optimizes the performance-efficiency trade-off, offering a scalable and practical solution for long-video understanding by selectively focusing perceptual resources on query-relevant segments.

\section{Conclusion}
\label{sec:conclusion}

This paper introduces \texttt{Video-QTR}, a lightweight, query-driven framework for long-video comprehension that overcomes the high computational cost and \textit{process-then-reason} limitation of existing models. \texttt{Video-QTR} utilizes a large-language model to determine \textit{what} and \textit{when} to perceive, incorporating three core components: the Reason Temporal Proxy (RTP) for temporal planning, a selective Perception Module for evidence gathering, and a Temporal Consistency Refiner (TCR) for causal consistency. Our experiments demonstrate that \texttt{Video-QTR} sets a new state-of-the-art across multiple benchmarks, achieving superior performance with significantly lower computational cost. Ablation studies highlight the critical role of RTP in driving this success. Notably, \texttt{Video-QTR} efficiently balances the accuracy-efficiency tradeoff, maintaining robustness even as video length increases. Looking ahead, \texttt{Video-QTR} opens avenues for research in hierarchical reasoning and interactive systems, advancing the development of scalable, practical video understanding systems.

{
    \small
    \bibliographystyle{ieeenat_fullname}
    \bibliography{main}
}
\clearpage

\clearpage
\maketitlesupplementary



\section{Detailed Performance Analysis}
\label{sec:appendix_a}

To substantiate the state-of-the-art performance reported in the main paper, this appendix provides a multi-faceted, in-depth analysis of our model's capabilities. We delve deeper than the aggregate scores to elucidate the precise sources of \texttt{Video-QTR}'s advantages. The analysis dissects our model's performance across two critical dimensions: (1) robustness to varying video durations on the demanding Video-MME benchmark, and (2) a fine-grained breakdown of its core reasoning and perceptual abilities. This granular examination is crucial for demonstrating not only \textit{that} our model performs well, but \textit{why} it succeeds where others falter.

\subsection{Performance Breakdown by Video Duration}

A primary challenge in long-video understanding is maintaining high accuracy as the temporal context expands and the amount of irrelevant information grows. Table~\ref{tab:duration_performance} provides compelling empirical evidence that \texttt{Video-QTR} not only meets this challenge but excels. While the main paper presents the overall scores, this detailed breakdown reveals a critical insight: \textbf{\texttt{Video-QTR}'s performance gap over competitors widens as video duration increases.}

For instance, in the challenging "Long (w/o subs)" category, our model achieves \textbf{66.46\%}, surpassing the next best single model, Gemini 2.5 Pro (61.49\%), by a significant margin of nearly 5 percentage points. This trend is even more pronounced in the "Long (w/ subs)" setting, where our model's \textbf{75.19\%} accuracy demonstrates a clear lead. This sustained high performance under temporal stress is directly attributable to our framework's core philosophy: by offloading high-level planning to the LLM and using a query-driven mechanism to invoke perception only when necessary, \texttt{Video-QTR} intelligently sidesteps the computational burden and cognitive distraction of processing redundant frames. This capability becomes an increasingly decisive advantage as video length grows, confirming the exceptional robustness and scalability of our approach.

\begin{table*}[!htbp]
    \centering
    \caption{Detailed performance comparison on the Video-MME benchmark, segmented by video duration. The evaluation is conducted under both "with subtitles" (w/ subs) and "without subtitles" (w/o subs) settings, revealing our model's strong performance, especially in long-video scenarios.}
    \label{tab:duration_performance}
    \resizebox{\textwidth}{!}{%
    \begin{tabular}{lcccccccc}
        \toprule
        \multirow{2}{*}{\textbf{Method}} & \multicolumn{2}{c}{\textbf{Short Acc. (\%)}} & \multicolumn{2}{c}{\textbf{Medium Acc. (\%)}} & \multicolumn{2}{c}{\textbf{Long Acc. (\%)}} & \multicolumn{2}{c}{\textbf{Overall Acc. (\%)}} \\
        \cmidrule(lr){2-3} \cmidrule(lr){4-5} \cmidrule(lr){6-7} \cmidrule(lr){8-9}
         & w/o subs & w/ subs & w/o subs & w/ subs & w/o subs & w/ subs & w/o subs & w/ subs \\
        \midrule
        \multicolumn{9}{l}{\textit{Comparison Methods}} \\
        Video-LLaVA & 45.30 & 46.10 & 29.00 & 40.70 & 36.20 & 29.10 & 39.90 & 41.60 \\
        ST-LLM & 45.70 & 48.00 & 36.80 & 41.00 & 31.30 & 36.90 & 37.90 & 42.30 \\
        VideoChat2-Mistral & 48.30 & 52.80 & 37.00 & 39.00 & 42.00 & 32.00 & 39.20 & 39.50 \\
        VITA-1.0 & 65.90 & 70.40 & 52.90 & 56.20 & 48.60 & 50.90 & 55.80 & 59.20 \\
        VILA-1.5 & 68.10 & 68.90 & 57.40 & 58.10 & 50.80 & 52.00 & 59.00 & 59.40 \\
        \midrule
        \multicolumn{9}{l}{\textit{Large Multimodal Models}} \\
        GPT-4.1 & 74.00 & 78.12 & 77.08 & 79.44 & 50.51 & 61.04 & 67.23 & 73.90 \\
        Gemini 2.5 Pro & 77.93 & 84.25 & 72.67 & 78.16 & 61.49 & 65.00 & 70.53 & 76.81 \\
        Qwen 2.5-VL-Max & 74.44 & 82.91 & 66.08 & 73.89 & 51.47 & 60.37 & 65.96 & 72.99 \\
        \midrule
        \textbf{Video-QTR (Ours)} & \textbf{73.91} & \textbf{83.33} & \textbf{76.44} & \textbf{84.27} & \textbf{66.46} & \textbf{75.19} & \textbf{72.59} & \textbf{80.93} \\
        \bottomrule
    \end{tabular}%
    }
\end{table*}

\newcommand{\cmark}{\textcolor{green!60!black}{\ding{51}}}%
\newcommand{\xmark}{\textcolor{red!80!black}{\ding{55}}}%

\begin{table*}[!htbp]
    \centering
    \caption{Detailed results of the ablation study on the Movie-Chat benchmark. We evaluate the impact of removing each core component: Reason-Temporal Proxy (RTP), Temporal Memory (TM), and Temporal Consistency Refiner (TCR). All metrics show a significant drop when any component is removed, with RTP being the most critical.}
    \label{tab:ablation_details}
    \resizebox{\textwidth}{!}{%
    \begin{tabular}{l ccc cc cc cc}
        \toprule
        \multirow{2}{*}{\textbf{Configuration}} & \textbf{RTP} & \textbf{TM} & \textbf{TCR} & \multicolumn{2}{c}{\textbf{Global Mode}} & \multicolumn{2}{c}{\textbf{Breakpoint Mode}} & \multicolumn{2}{c}{\textbf{Total}} \\
        \cmidrule(lr){5-6} \cmidrule(lr){7-8} \cmidrule(lr){9-10}
         & (Query-driven) & (Graph Memory) & (Refinement) & \textbf{Acc. (\%)} & \textbf{Score} & \textbf{Acc. (\%)} & \textbf{Score} & \textbf{Acc. (\%)} & \textbf{Score} \\
        \midrule
        \multicolumn{10}{l}{\textit{Ablation of individual components from our full model:}} \\
        w/o RTP (uses random selection) & \xmark & \cmark & \cmark & 69.44 & 3.49 & 40.98 & 2.06 & 47.55 & 2.38 \\
        w/o TM & \cmark & \xmark & \cmark & 79.70 & 3.98 & 67.17 & 3.36 & 71.60 & 3.58 \\
        w/o TCR & \cmark & \cmark & \xmark & 73.91 & 3.69 & 61.94 & 3.11 & 64.70 & 3.24 \\
        \midrule
        \textbf{Video-QTR (Full Model)} & \cmark & \cmark & \cmark & \textbf{88.72} & \textbf{4.45} & \textbf{74.72} & \textbf{3.60} & \textbf{77.95} & \textbf{3.94} \\
        \bottomrule
    \end{tabular}%
    }
\end{table*}

\subsection{Fine-grained Capability Analysis}
\begin{figure}[h!]
    \centering
    \includegraphics[width=0.8\linewidth]{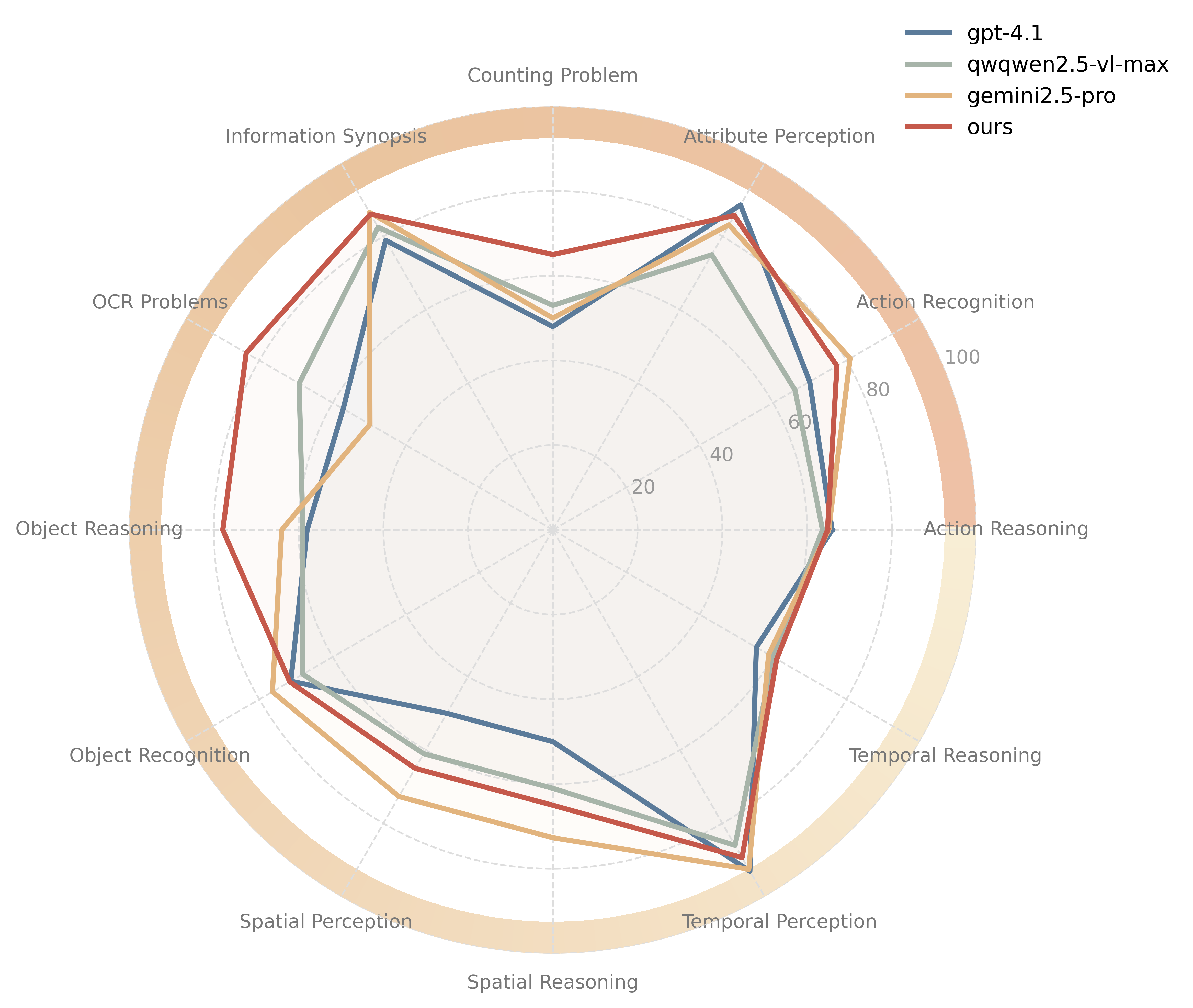} 
    \caption{\textbf{Fine-grained capability analysis on the Video-MME benchmark.} This radar chart provides a detailed breakdown of performance across 12 distinct reasoning and perception abilities, comparing Video-QTR against leading MLLMs. The larger area covered by our model (red) indicates superior and more balanced capabilities, especially in temporal and action-related reasoning tasks.}
    \label{fig:capability_radar}
\end{figure}

Furthermore, to understand the source of Video-QTR's strong performance, we present a fine-grained capability analysis in Figure~\ref{fig:capability_radar}. The chart visualizes performance across the 12 evaluation dimensions of the Video-MME benchmark. Notably, Video-QTR (represented by the red area) exhibits a significant advantage in higher-order reasoning tasks such as \textbf{Temporal Reasoning}, \textbf{Action Reasoning}, and \textbf{Causal Reasoning}. This empirically validates that our query-driven framework, which iteratively refines its understanding, effectively enhances the model's ability to comprehend complex temporal dynamics and event relationships, distinguishing it from models that rely on more passive, exhaustive frame processing.

\section{Ablation Study Details}
\label{sec:appendix_b}

To rigorously validate the architectural design of \texttt{Video-QTR} and to quantify the synergistic contribution of each of its core components, we conducted a comprehensive ablation study on the challenging Movie-Chat benchmark. This section provides the detailed quantitative results that form the empirical foundation for the summary visualization presented in Figure 3 of the main paper. As shown in Table~\ref{tab:ablation_details}, the removal of any single component leads to a notable degradation in performance, which underscores their interdependent roles. We analyze the impact of each below:

\begin{itemize}
    \item \textbf{Reason-Temporal Proxy (RTP):} The removal of the RTP module results in the most substantial performance degradation across all metrics. In this configuration, the model loses its query-driven guidance and defaults to a less-informed strategy (e.g., random segment selection). This sharp decline empirically validates our central hypothesis: a "reason-then-perceive" approach, where perception is actively guided by semantic intent, is critically superior to directionless exploration. The RTP is the linchpin of this entire paradigm.

    \item \textbf{Temporal Memory (TM):} Disabling the Temporal Memory, which aggregates findings across iterations, leads to a significant performance drop, particularly in tasks requiring long-horizon contextual understanding. Without the TM, the model effectively becomes "stateless" between reasoning steps, unable to build upon past observations or infer relationships between events that occur far apart in time. The TM's contribution is therefore vital for achieving robust, cross-scene comprehension.

    \item \textbf{Temporal Consistency Refiner (TCR):} The absence of the TCR, our mechanism for grounding symbolic reasoning in temporal reality, also causes a marked decrease in accuracy. This demonstrates that the LLM's initial temporal plan, while generally effective, can sometimes diverge from the video's actual chronological flow. The TCR acts as an essential corrective feedback loop, ensuring that the model's evolving understanding remains causally and temporally coherent, thereby preventing reasoning errors stemming from misaligned event sequences.
\end{itemize}

In summary, the detailed data in Table~\ref{tab:ablation_details} not only quantifies the individual importance of the RTP, TM, and TCR but also collectively illustrates that the high performance of \texttt{Video-QTR} is an emergent property of their sophisticated interplay.

\begin{figure*}[htbp!]
\centering
\begin{tcolorbox}[
    colback=black!5!white,
    colframe=black!75!black,
    title=Prompt Template for Reason-Temporal Proxy (RTP),
    fonttitle=\bfseries,
    width=\textwidth,
    fontupper=\ttfamily 
]
\small
\textbf{SYSTEM PROMPT:}\\[0.5em]
You are an intelligent video analyst. Your primary goal is to determine the most relevant time segment in a video to answer a user's question. Please avoid selecting timeframes that have already been reviewed. Here are the segments already analyzed: \textit{\{list of previously reviewed timeframes\}}

\vspace{1em}
\textbf{USER PROMPT:}\\[0.5em]
I will provide you with a question and the total duration of a video. Based on this information, you need to return the \textbf{single time segment} you believe is most crucial for answering the question.

\textbf{Context:}\\[0.5em]
- \textbf{Question:} "\textit{\{formatted\_question with options\}}"\\
- \textbf{Total Video Duration:} \textit{\{video\_duration\}} seconds

\vspace{1em}
\textbf{Task:}\\[0.5em]
1.  Analyze the question and any historical context from previously reviewed segments.\\[0.5em]
2.  Decide which part of the video you need to see \textbf{next} to make progress on the answer.\\[0.5em]
3.  Return only the start and end times of this single segment.

\vspace{1em}
\textbf{Constraints (You MUST follow these):}\\[0.5em]
1.  The output format must be a JSON array: \textbf{[start\_time, end\_time]}.\\[0.5em]
2.  The duration of your selected segment must NOT exceed 180 seconds (i.e., end\_time - start\_time <= 180).\\[0.5em]
3.  The \textbf{end\_time} must be greater than the \textbf{start\_time}.\\[0.5em]
4.  Do NOT select the entire video (e.g., [0, \textit{\{video\_duration\}}]).\\[0.5em]
5.  Do NOT select segments that have been previously reviewed.

\vspace{1em}
Return ONLY the JSON array and nothing else.
\end{tcolorbox}
\caption{The detailed prompt template for the RTP module. This prompt instructs the LLM to act as a strategic agent, iteratively selecting the most informative temporal windows for analysis while adhering to strict formatting and duration constraints.}
\label{fig:rtp_prompt_template}
\end{figure*}


\section{Prompting Details for Core Modules}
\label{sec:appendix_prompts}

At the heart of the \texttt{Video-QTR} framework lies a sophisticated prompt engineering strategy, designed to harness the advanced reasoning capabilities of Large Language Models (LLMs) while enforcing strict operational constraints. This section elucidates the specific prompts that orchestrate the interaction between the LLM—serving as the cognitive engine—and the framework's other modules. The meticulously engineered structure of these prompts is instrumental in transforming the LLM from a general-purpose model into a specialized, predictable, and controllable component of our query-driven pipeline, making our entire approach both effective and reproducible.

\subsection{Prompt for Reason-Temporal Proxy (RTP)}
\label{sec:appendix_rtp_prompt}

The Reason-Temporal Proxy (RTP) spearheads the paradigm shift from passive processing to active, query-driven reasoning. It orchestrates the initial and most critical phase of our iterative loop: strategic temporal planning. Instead of brute-force frame analysis, the RTP module intelligently queries the LLM to hypothesize where the most salient information might lie.

The prompt detailed in Figure~\ref{fig:rtp_prompt_template} is precisely engineered to compel the LLM to function as a strategic planner. It furnishes the model with a complete state-of-the-task overview, including the user's query, the total video duration, and, crucially, a memory of previously analyzed segments to prevent redundant exploration. In response, the LLM is constrained to output a single, well-defined time segment, ensuring that its proposal is not only semantically relevant but also computationally tractable. This mechanism is the direct implementation of our "reason-then-perceive" philosophy, forming the foundation for the framework's efficiency and high performance.

\begin{figure*}[htbp!]
\centering
\begin{tcolorbox}[
    colback=black!5!white,
    colframe=black!75!black,
    title=Prompt Template for Answering Agent,
    fonttitle=\bfseries,
    width=\textwidth,
    fontupper=\ttfamily
]
\small
\textbf{SYSTEM PROMPT:}\\[0.5em]
You are an expert video analyst. You will be given a video clip and a question. If there is a history record, you can refer to it and the content from clips you selected before to assist you. Your goal is to answer the user's question based \textbf{only} on the information present in the provided video clips.

Your answer must include a 'Confidence Score' that reflects how completely the provided evidence answers the question. You MUST adhere to the scoring guide below.

\vspace{1em}
\textbf{Confidence Score Guide:}\\[0.5em]
- \textbf{High Confidence (90-100):} The video clip(s) provide a complete, unambiguous, and definitive answer. No other information is needed.\\[0.5em]
- \textbf{Medium Confidence (40-89):} The clip(s) provide a partial answer, a strong hint, or an answer that is not fully clear. More details might exist elsewhere in the video.\\[0.5em]
- \textbf{Low Confidence (1-39):} The clip(s) are irrelevant or provide virtually no useful information to answer the question.

\vspace{1em}
\textbf{USER PROMPT:}\\[0.5em]
- \textbf{Question:} "\textit{\{formatted\_question with options\}}"\\
- \textbf{Video Clip:} \textit{\{A video clip corresponding to a specific time segment is provided here.\}}

\vspace{1em}
\textbf{Task:}\\[0.5em]
Analyze the video clip and answer the question by strictly following the output format below.

\vspace{1em}
\textbf{Output Format (You MUST follow this format EXACTLY):}\\[0.5em]
Answer: <Your answer based ONLY on the evidence from the clip(s)>\\
Reason: <Explain precisely how the content of the clip(s) leads to your answer and confidence level>\\
Summary of this content: <A brief, objective summary of the events in the latest video clip>\\
Confidence Score: <A score from 1-100 based on the guide above>
\end{tcolorbox}
\caption{The detailed prompt template for the Answering Agent. This prompt compels the model to ground its response in visual evidence and to self-evaluate its confidence, which is a key mechanism for enabling iterative reasoning and stopping criteria.}
\label{fig:answer_prompt_template}
\end{figure*}

\subsection{Prompt for Answering Agent}
\label{sec:appendix_answer_prompt}

Following the targeted perception phase guided by the RTP, the Answering Agent performs the critical task of evidence synthesis and evaluation. This module serves as the evaluative counterpart to the RTP's exploratory function, closing the "reason-then-perceive" loop. The prompt detailed in Figure~\ref{fig:answer_prompt_template} is therefore not merely a request for an answer; it is a structured interrogation designed to compel the LLM to adopt the role of a cautious and meticulous analyst.

The most pivotal element of this prompt is the mandatory inclusion of a \textbf{Confidence Score}, governed by a strict, predefined rubric. This mechanism serves two fundamental purposes that are indispensable to our framework's success:

\begin{enumerate}
    \item \textbf{A Feedback Signal for Iteration:} The Confidence Score acts as the primary feedback signal to the entire control loop. A low or medium score quantitatively communicates to the framework that the current evidence is insufficient, thereby triggering another iteration of the RTP's temporal planning to seek out missing information.
    \item \textbf{A Robust Stopping Criterion:} Conversely, a high confidence score provides a reliable, data-driven criterion for terminating the iterative search. This ensures that the framework ceases exploration once a definitive answer is found, optimizing the trade-off between answer completeness and computational expenditure.
\end{enumerate}

By forcing the LLM to perform this quantitative self-assessment—distinguishing between definitive proof, partial clues, and irrelevant noise—this mechanism directly counteracts the well-documented tendencies of MLLMs toward hallucination and overconfidence. It is this capacity for self-aware reasoning that underpins \texttt{Video-QTR}'s ability to handle ambiguity and deliver trustworthy, evidence-grounded answers.

\begin{figure*}[htbp]
    \centering
    \includegraphics[width=\textwidth]{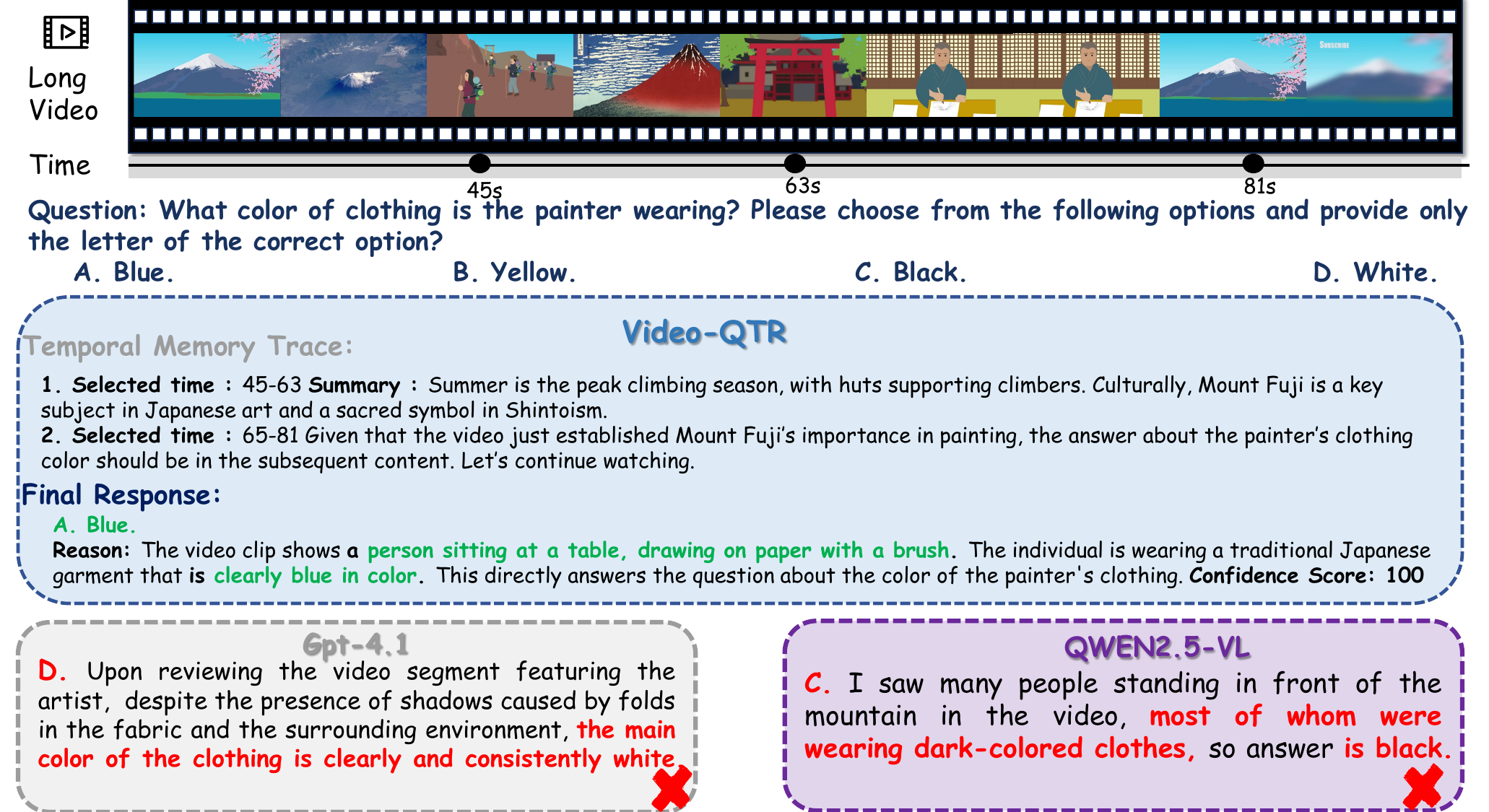} 
    \caption{\textbf{Qualitative comparison on a short-video perception task.} This example highlights \texttt{Video-QTR}'s robustness against common failure modes. Through iterative refinement shown in its \textit{Temporal Memory Trace}, our model correctly identifies the painter's blue clothing. In contrast, other leading models succumb to either perceptual errors induced by lighting (\texttt{GPT-4.1}) or severe contextual hallucination (\texttt{Qwen-VL-Max}).}
    \label{fig:case_short}
\end{figure*}

\section{Qualitative Case Studies}
\label{sec:appendix_cases}

To complement the quantitative results presented in our main paper, this section provides a series of qualitative case studies. These examples serve to elucidate the practical implications of our \texttt{Video-QTR} framework, offering a transparent view into its iterative reasoning process. Through direct juxtaposition with other leading models like GPT-4.1 and Qwen-VL-Max, we demonstrate not only \textit{that} our model achieves superior accuracy, but also \textit{how} it successfully navigates challenges such as perceptual ambiguity, contextual distraction, and long-range temporal reasoning, where other models often falter.

\subsection{Case 1: Overcoming Perpetual Ambiguity and Distraction}

Figure \ref{fig:case_short} illustrates a scenario that tests a model's ability to perform accurate visual perception amidst potential distractions. The question is straightforward: identifying the color of a painter's clothing. However, this seemingly simple task proves challenging for leading models.

As shown in the comparison, \texttt{GPT-4.1} incorrectly identifies the clothing as "white," likely due to being misled by lighting effects or shadows on the fabric. \texttt{Qwen-VL-Max} makes an even more severe error, defaulting to a contextual hallucination by guessing "black" based on an incorrect premise that "most people... were wearing dark-colored clothes." This demonstrates a failure to ground its response in the specific visual evidence related to the question's subject.

In stark contrast, \texttt{Video-QTR}'s \textit{Temporal Memory Trace} reveals a methodical, two-step reasoning process. It first identifies the general context (Mount Fuji's importance in art) and then correctly deduces that the answer lies in the subsequent segment focusing on the painter. This query-driven refinement allows it to precisely locate the correct visual evidence and confidently identify the color as "Blue," avoiding both perceptual errors and contextual distractions.

\begin{figure*}[htbp]
    \centering
    \includegraphics[width=\textwidth]{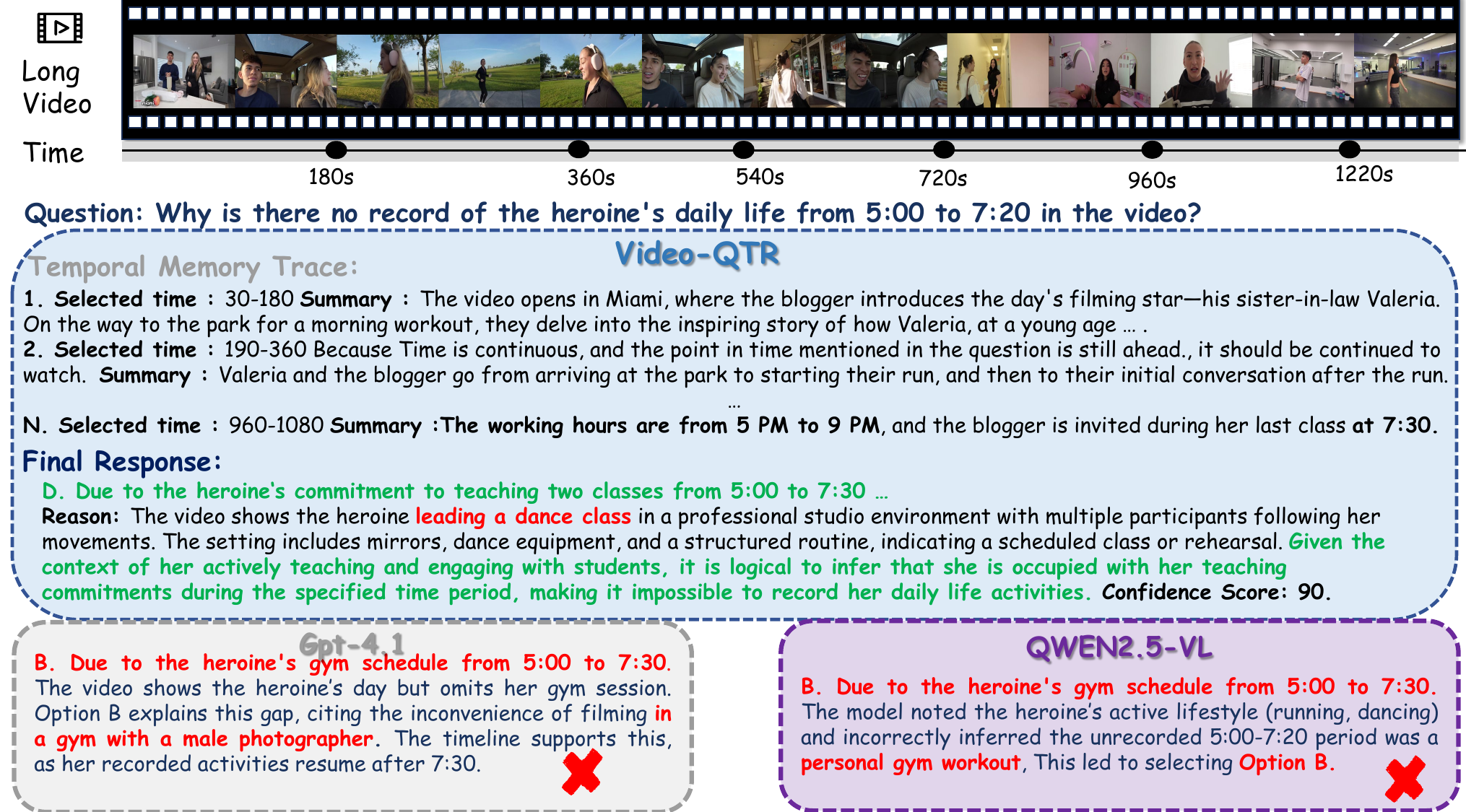} 
    \caption{\textbf{Qualitative comparison on a long-video reasoning task.} This case demonstrates \texttt{Video-QTR}'s superior ability in long-range temporal reasoning. It successfully answers a complex "why" question by synthesizing evidence from disparate time segments (an early context and a late-video clue). In contrast, both \texttt{GPT-4.1} and \texttt{Qwen-VL-Max} fail by generating plausible but unfounded inferences, highlighting their limitations in maintaining and querying context over extended durations.}
    \label{fig:case_long}
\end{figure*}

\subsection{Case 2: Long-Range Temporal Reasoning and Inference}

The example in Figure \ref{fig:case_long} presents a more complex challenge inherent to long-video understanding: answering a question that requires synthesizing information from non-adjacent segments of a lengthy video. The question asks for the reason behind a gap in the protagonist's recorded daily life.

Both \texttt{GPT-4.1} and \texttt{Qwen-VL-Max} fail by converging on a plausible but factually incorrect hallucination: that the heroine was at the gym. This represents a classic failure mode where models, lacking a robust mechanism for long-range evidence gathering, default to commonsense inferences that are not grounded in the source material.

\texttt{Video-QTR}, however, showcases the power of its iterative search and memory capabilities. Its \textit{Temporal Memory Trace} documents a journey through the video: after analyzing early segments and realizing the answer is not present, it continues its exploration. It eventually discovers the crucial piece of evidence much later in the video (at the 960-1080s mark), which reveals the heroine's teaching schedule. By retaining and building upon its memory across multiple "reason-then-perceive" cycles, our model correctly pieces together the disparate clues to deduce the correct reason for the time gap. This case vividly illustrates how our framework's architecture directly addresses the core challenges of long-horizon temporal reasoning.

\section{Broader Impacts}
\label{sec:appendix_broader_impacts}

The introduction of \texttt{Video-QTR} presents a paradigm shift towards more efficient video understanding, which carries both significant positive potential and notable risks that warrant careful consideration. As researchers, we acknowledge our responsibility to discuss these broader implications.

\subsection{Positive Societal Impacts}

The core contribution of our work lies in its lightweight and query-driven nature, which could lead to several positive societal outcomes:

\begin{itemize}
    \item \textbf{Democratizing Advanced Video Analysis:} By drastically reducing the computational resources required for long-video understanding (up to 73\% fewer frames processed), \texttt{Video-QTR} lowers the barrier to entry. This enables smaller research labs, startups, and even individuals without access to massive GPU clusters to develop and deploy sophisticated video analysis tools, fostering broader innovation.

    \item \textbf{Promoting Sustainable AI ("Green AI"):} The significant reduction in computational load directly translates to lower energy consumption. As the scale of AI models continues to grow, frameworks like ours that prioritize efficiency contribute to a more environmentally sustainable research and deployment ecosystem.

    \item \textbf{Enabling New Assistive Technologies:} The ability to efficiently query long videos in near real-time opens doors for novel assistive applications. For example, it could power tools that help visually impaired users quickly understand the contents of a long, uncurated video, or create automated summaries of lengthy educational lectures or public meetings, enhancing accessibility to information.
\end{itemize}

\subsection{Potential Risks and Ethical Considerations}

Like any powerful analysis tool, \texttt{Video-QTR} has a dual-use nature, and its misuse could lead to negative consequences:

\begin{itemize}
    \item \textbf{Propagation of Inherited Biases:} The framework's reasoning core is an LLM. Therefore, any societal biases (e.g., racial, gender, or cultural) present in the pre-trained LLM could be propagated and even amplified. For instance, a biased LLM might generate temporal reasoning plans that preferentially focus on certain demographics, leading to skewed or unfair outcomes in its analysis.

    \item \textbf{Misuse for Large-Scale Surveillance:} The high efficiency and scalability of our framework could make large-scale, automated surveillance more affordable and easier to implement. If deployed without strict ethical oversight and legal regulation, such technology could be used to monitor public or private spaces in ways that infringe upon individual privacy and civil liberties.

    \item \textbf{Automated Generation of Misinformation:} The ability to quickly extract specific segments from long videos based on a query could be weaponized to create misleading narratives or "fake news." Malicious actors could use such a tool to cherry-pick moments out of context to support a false claim, thereby automating the production of convincing propaganda.
\end{itemize}

We believe that the development of efficient AI systems like \texttt{Video-QTR} must proceed in parallel with the establishment of strong ethical guardrails, transparency standards, and robust public discourse on their appropriate use.

\section{Limitations}
\label{sec:appendix_limitations}

Despite its strong performance, we acknowledge several limitations in our \texttt{Video-QTR} framework that suggest avenues for future research:

\begin{itemize}
    \item \textbf{Dependence on Foundational LLM:} The framework's performance is fundamentally bound by the reasoning quality of the underlying LLM, whose errors or biases can propagate into the temporal planning stage.

    \item \textbf{Challenges in Perception Strategy:} The perception strategy faces a "cold start" problem for non-obvious queries and risks overlooking brief but critical events due to its segment-based nature.
\end{itemize}

Future work could explore hybrid perception strategies or lightweight preview mechanisms to mitigate these challenges.


\end{document}